\documentclass{article}
\usepackage{microtype}
\usepackage{graphicx}
\usepackage{booktabs} 

\usepackage{amsmath,amsfonts,bm}

\def\eqref#1{equation~\ref{#1}}

\def\1{\bm{1}}

\DeclareMathAlphabet{\mathsfit}{\encodingdefault}{\sfdefault}{m}{sl}
\SetMathAlphabet{\mathsfit}{bold}{\encodingdefault}{\sfdefault}{bx}{n}

\usepackage{url}
\usepackage{algorithm}
\usepackage{algorithmic}
\usepackage{graphicx}
\usepackage{amsthm}
\usepackage{subfig}
\usepackage{hyperref}

\newtheorem{definition}{Definition}
\newcommand{\begd}{\begin{definition}}
\newcommand{\enf}{\end{definition}}
\newtheorem{theorem}{Theorem}
\newenvironment{sproof}{%
 \proof}{\endproof}

\usepackage[accepted]{icml2019}

\begin{document}


\twocolumn[
\icmltitle{Privacy Preserving Adjacency Spectral Embedding on Stochastic Blockmodels}


\icmlsetsymbol{equal}{*}

\begin{icmlauthorlist}
\icmlauthor{Li Chen}{equal,to}

\end{icmlauthorlist}

\icmlaffiliation{to}{Security and Privacy Research, Intel Labs}

\icmlcorrespondingauthor{Li Chen}{li.chen@intel.com}

\icmlkeywords{Machine Learning, ICML}

\vskip 0.3in
]



\printAffiliationsAndNotice{}  

\begin{abstract}
For graphs generated from stochastic blockmodels, adjacency spectral embedding is asymptotically consistent. Further, adjacency spectral embedding composed with universally consistent classifiers is universally consistent to achieve the Bayes error. However when the graph contains private or sensitive information, treating the data as non-private can potentially leak privacy and incur disclosure risks. In this paper, we propose a differentially private adjacency spectral embedding algorithm for stochastic blockmodels. We demonstrate that our proposed methodology can estimate the latent positions close to, in Frobenius norm, the latent positions by adjacency spectral embedding and achieve comparable accuracy at desired privacy parameters in simulated and real world networks. 
\end{abstract}

\section{Introduction}


Our framework considers random graphs. A random graph is a graph-valued random variable, with a fixed vertex set and a random edge set generated from a probabilistic distribution. The latent position models \citep{hoff2002latent}, random dot product graphs \citep{young2007random} and stochastic blockmodels \citep{sbm, sbm2} are a few examples of random graphs. Many algorithms develop ed on random graphs indicate vertices from the same class share similar connectivity patterns.  One  important exploitation  task  on  such datasets is  vertex  classification to identify  the class labels of the vertices. For example, we may wish to classify whether a user in a social network holds liberal or conservative political views, or whether an individual in a communication network belongs to a social community.

The adjacency spectral embedding and Laplacian spectral embedding methods are valuable for performing inference on graphs realized from a stochastic blockmodels \citep{sussman2014consistent, fishkind2013consistent, stfp, tang2013universally}. Theoretically they are proven to be consistent estimation \citep{devroye1996probabilistic} of the latent positions that generate the stochastic blockmodels. Classification techniques such as $k$-nearest neighbor composed with spectral embedding are proved to be Bayes optimal, which indicate such composition can theoretically obtain the minimum misclassification error. 

On the other hand, graphs that describe social networks, communication networks or purchasing networks can contain sensitive and private information about individuals such as their communication frequencies, patterns, webpage visits, purchasing behaviors and so on. It becomes critical to propose and design inference algorithms on random graphs for not only generating effective and accurate analysis results but also protecting the sensitive nature of the information. 

Differential privacy \citep{dwork2011differential} is a mechanism that measures privacy risk via parameters $\alpha$ or ($\alpha, \delta$) and bounds the likelihood of the algorithm when two data sets differ by one sample. A randomized algorithm is considered differentially private when its outputs cannot expose whether a particular individual's information was used in the computation. Essentially, differential privacy requires the probability distribution on the analysis results, over the outcome space of the underlying algorithm, to be roughly the same, independent of whether or not any individual exists in the data set.

Motivated by the practical challenges faced with inference on random graphs containing sensitive information, in this paper, we propose a differentially private adjacency spectral embedding procedure on stochastic blockmodels to estimate the latent positions and further composite it with universally consistent classifier such as $k$ nearest neighbor for vertex classification. We demonstrate in our experiments the classification efficacy in simulated experiments and two real-world networks.

The rest of the paper is organized as follows. In Section 2, we provide brief background on graph models and differential privacy. In Section 3, we present our proposed algorithm differentially private adjacency spectral embedding. In Section 4, we present results on numerical simulation, a political blog network and a co-purchasing product network. 


\section{Preliminaries}
\subsection{Graph Models}

A random graph is a graph-valued random variable: $\mathcal{G}: \Omega \rightarrow \mathcal{G}_n$, where $\mathcal{G}_n$ represents the collection of all $2^{\binom{[n]}{2}}$ possible graphs on the vertex set $V = [n]$, and $\Omega$ is a probability space. Associated with the adjacency matrix $A \in \mathbb{R}^{n \times n}$, there exists a probability matrix $P \in [0,1]^{n \times n}$, where each entry $P_{ij}$ denotes the probability of edge between vertex $i$ and vertex $j$. 

In the latent position model (LPM) \citep{hoff2002latent}, each vertex $i$ is associated with a latent random variable $X_i \in \mathbb{R}^d$ drawn independently from a specified distribution $F$ on $\mathbb{R}^{d}$. These latent variables determine the probabilities of edge existence. The adjacency matrix entries $A_{ij}|(X_{i}, X_{j}) \sim \text{Bernoulli}(l(X_{i}, X_{j}))$ are conditionally independent, where $l: \mathbb{R}^{d} \times \mathbb{R}^{d} \rightarrow [0,1]$ is defined as the link function. 

The stochastic blockmodel (SBM) is a special case of LPM. The latent positions of an SBM are sampled as a mixture of the point masses which are the eigenvectors of $B$. The SBM \citep{holland1983stochastic} is a family of random graph models with a set of $n$ vertices randomly belonging to $K$ blocks. Conditioned on the $K$-partition, edges between all the pairs of vertices are independent Bernoulli trials with parameters determined by the block memberships of the two vertices. 

\begin{definition} \label{def:sbm1} \textbf{Stochastic Blockmodel SBM$([n], B, \pi)$} 
\normalfont Let $K$ be the number of blocks. Let $\pi$ be a length $K$ vector in the unit simplex $\Delta^{K-1}$ specifying the block membership probabilities. The block membership of the vertex $i$ is given by $Y_i \overset{iid}\sim \text{Multinomial}([K], \pi)$. Let $B$ be a $K \times K$ symmetric block communication probability matrix. Then the graph $G$ is realized from a stochastic blockmodel $G \sim SBM([n], B, \pi)$ if 
\begin{equation}
\mathbb{P}(A | Y_{1}, ..., Y_{n}) = \Pi_{i < j}P_{ij}^{A_{ij}}(1-P_{ij})^{1-A_{ij}}, 
\end{equation}
\begin{equation}
P_{ij} = \mathbb{P}(A_{ij} =1| X_{i}, X_{j}) = \mathbb{P}(A_{ij} =1| Y_{i}, Y_{j}) = B_{Y_{i}, Y_{j}}.
\end{equation}
\end{definition}
 
\subsection{Adjacency Spectral Embedding}\label{sec:models_ase}
Adjacency spectral embedding (ASE)\citep{stfp} has theoretical guarantee to consistently estimate the latent positions of SBM. This technique applies spectral decomposition on $A$ to compute the first $d$ eigen-pairs $(U_A, S_A) \in \mathbb{R}^{n\times d} \times \mathbb{R}^d$, where $S_A$ is diagonal with the top $d$ eigenvalues sorted in absolute values. The resulted embedding $ASE := U_AS_A^{\frac{1}{2}}$ consistently estimates the block memberships of SBM. The consistency property of Laplacian spectral embedding is proved in \cite{rohe2011spectral}.

Adjacency spectral embedding followed by $k$-nearest neighbor ($k$NN) classifiers \citep{sussman2014consistent}, linear classifiers\citep{tang2013universally} is shown to be universally consistent, which means such procedure achieves the minimum Bayes error as $n$ goes to infinity.

\subsection{Differential Privacy}
Denote a randomized algorithm by $f$ taking values in a set $S$. $A$ and $B$ are datasets that differ by one element. 

\begin{definition}\textbf{$\alpha$-Differential Privacy}
$f$ provides $\alpha$-differential privacy if 
\begin{equation}
    \mathbb{P}(f(A) \in S) \leq \mathbb{P}(f(B) \in S) e^{\alpha}.
\end{equation}
\end{definition}

\begin{definition}\textbf{($\alpha$, $\delta$)-Differential Privacy}
$f$ provides ($\alpha$, $\delta$)-differential privacy if 
\begin{equation}
    \mathbb{P}(f(A) \in S) \leq \mathbb{P}(f(B) \in S) e^{\alpha} + \delta.
\end{equation}
\end{definition}

The parameter $\alpha$ bounds on the change in probability of any outcome. A low value of $\alpha$ such as 0.1 implies that very little can change in the beliefs about any sample's existence in the database. The ($\alpha$, $\delta$)-Differential Privacy is a weaker notion compared with the $\alpha$-Differential Privacy. 
\section{Differentially Private Adjacency Spectral Embedding}
We propose a differentially private mechanism for adjacency spectral embedding (DP-ASE) to estimate the latent positions of stochastic blockmodels (SBM).

\begin{algorithm}[H]
\caption{DP-ASE for SBM}
\begin{algorithmic}
\STATE \textbf{Input: } The adjacency matrix $A$. The embedding dimension $d$.  Desired privacy parameters $(\alpha, \delta)$.\\
\STATE \textbf{Output: } Differentially private approximations of latent positions $\mathcal{X}_{DP} \in \mathbb{R}^{n \times d}$.
\STATE \textbf{Step 1:} Set $\beta^2 := \frac{8d^2\log^2(d/\delta)}{n^2\alpha^2}$. Sample a matrix $E$ whose entries are i.i.d from normal distribution $\mathcal{N}(0, \beta^2)$. Symmetrize $E$.
\STATE \textbf{Step 2:} Apply spectral embedding on $A_{DP} := A + E$. Compute the first $d$ eigen-pairs of $A_{DP}$, denoted by $(U_{DP}, S_{DP})\in \mathbb{R}^{n \times d} \times \mathbb{R}^{d}$, where $S_{DP}$ has $d$ largest eigenvalues in magnitude sorted in non-increasing order.
\STATE \textbf{Step 3:} Denote the $d$-dimensional coordinate-scaled singular vector matrix of $A_{DP}$ to be $\mathcal{X}_{DP} := U_{DP}S_{DP}^{1/2} \in \mathbb{R}^{n \times d}$.  
\end{algorithmic}
\label{alg: dp_ase}
\end{algorithm}

\begin{theorem}[Privacy of DP-ASE]
Algorithm \ref{alg: dp_ase} computes an $(\alpha, \delta)$-differentially private approximation to adjacency spectral embedding on stochastic blockmodels.
\label{thm:dp-ase}
\end{theorem}
\begin{sproof}
The proof of Theorem \ref{thm:dp-ase} follows directly from the differentially private algorithm SULQ to approximate PCA with privacy guarantee \citep{blum2005practical, chaudhuri2012near}.  
\end{sproof}
\section{Experiments}
An effective differentially private algorithm should not only provide privacy guarantee but also generate reasonably accurate results. In our experiments, we compose a universally consistent classifier $k$ nearest neighbor with DP-ASE, denoted by $k$NN$\circ$DP-ASE and evaluate its classification error. The exploitation task is to correctly predict the block memberships of the vertices. The comparison baseline is the classification error by $k$NN$\circ$ASE. We set $k = 3$ in the nearest neighbor classifier, use leave-one-out cross validation to estimate the classification error and generate new matrix $E \sim \mathcal{N}(0, \beta^2)$ in each Monte Carlo replicate. 
\subsection{Simulation}
We simulate SBM with $K=2$ blocks ($Y \in \{1,2\}$) and parameters
$
  B =
\left( {\begin{array}{cc}
 0.3 & 0.1 \\
 0.1 & 0.2
 \end{array} } \right)  
$, $\pi=[0.4, \enspace 0.6]^{T}$.

\begin{figure*}[t]
\includegraphics[width=0.26\textwidth]{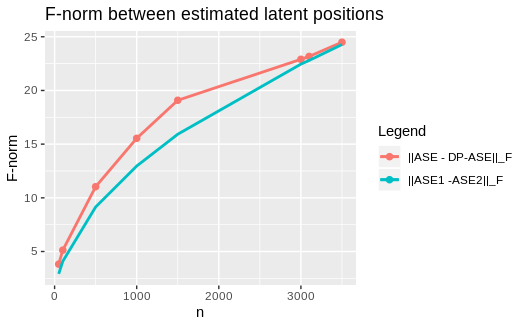}
\includegraphics[width=0.26\textwidth]{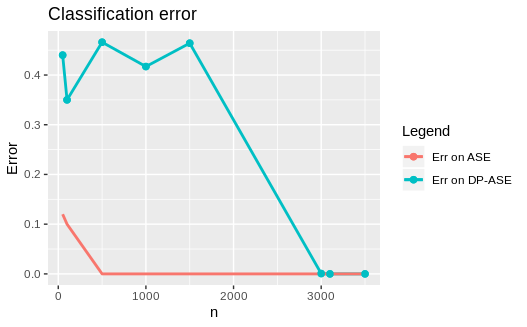}
\includegraphics[width=0.18\textwidth]{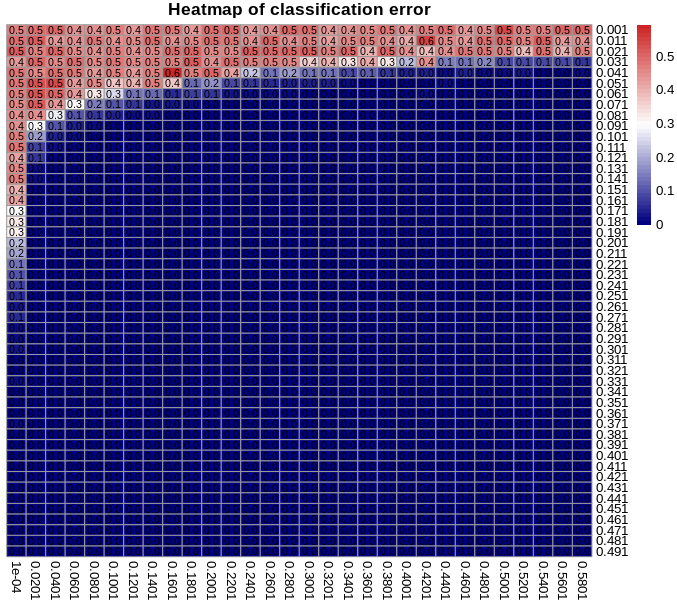}
\includegraphics[width=0.18\textwidth]{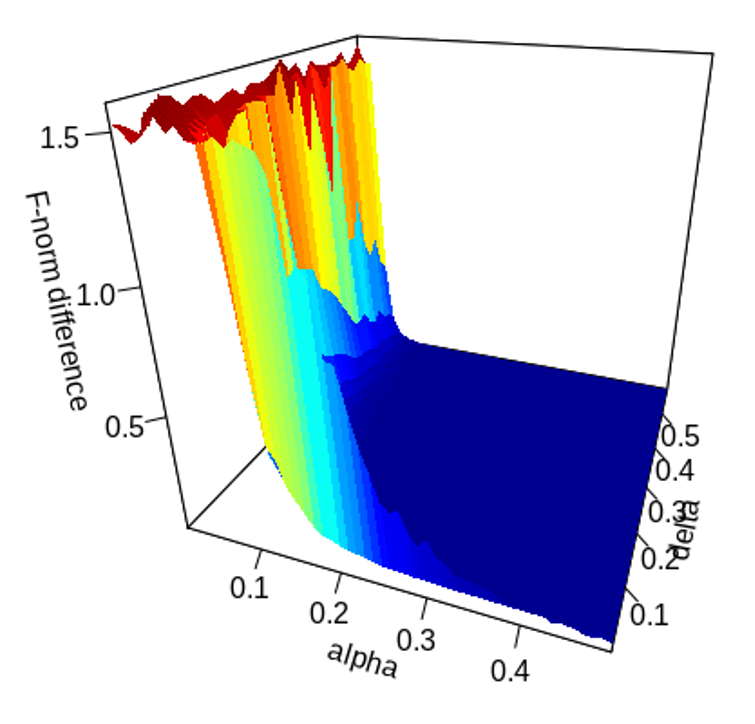}
\caption{\label{fig:sim}(1) $F$-norm of the estimated latent positions. As $n$ increases, the distance of the estimated latent positions via ASE and DP-ASE become closer. (2) Classification error. When $n$ is small, $k$NN$\circ$DP-ASE has higher classification error than $k$NN$\circ$ASE. As $n$ increases, classification error by $k$NN$\circ$ goes to the Bayes error at zero. (3) 3D plot of $F$-norm difference as $\alpha$ and $\delta$ vary. (4) Heat map of classification error as $\alpha$ and $\delta$ vary. $k$NN$\circ$DP-ASE achieves Bayes error at privacy parameters with lower values.}
\end{figure*}

Our first experiment is to compare the latent position approximation and classification performance with or without preserving privacy at fixed but low privacy budget, as the number of vertices increases. 
We apply ASE and DP-ASE at fixed privacy parameters $\alpha = 0.1$ and $\delta = 0.001$. We first vary the number of vertices $n \in \{50, 100, 500, 1000, 1500, 3000, 3100, 3500, 4000 \}$ and compare two terms: 1. the Frobenius norm (F-norm) of the latent positions estimated by DP-ASE and ASE respectively, and the F-norm of the latent positions estimated by two ASEs aligned via solving the orthogonal Procrustes fit problem \cite{gower2004procrustes}; 2.
the classification errors by $k$NN$\circ$DP-ASE and $k$NN$\circ$ASE respectively at each $n$. As $n$ increases, DP-ASE approximates to ASE in both latent position estimation and classification error, as seen in left two figures in Fig \ref{fig:sim}. 

Next we understand the utility-privacy tradeoff for smaller-sized stochastic blockmodels. We set $n = 300$, vary the privacy parameters $\alpha \in \{0.001, 0.011, 0.021,...,0.05 \}$ and $\delta \in \{0.0001, 0.0021, 0.0041, ..., 0.6 \}$ and examine the classification error of $k$NN$\circ$DP-ASE. We inspect the minimum levels of privacy parameters for $k$NN$\circ$DP-ASE to achieve the Bayes error, which is zero in this simulation. In the heat map of Fig \ref{fig:sim}, the rows are $\alpha$ and columns are $\delta$. $k$NN$\circ$DP-ASE achieves the Bayes error at the lower values of the privacy parameters, which provide tighter privacy bounds.

\subsection{Real Data Experiments}
\subsubsection{Political Blog Sphere}
\begin{figure}
\centering
\centering
\subfloat[]{
\includegraphics[width=0.26\textwidth]{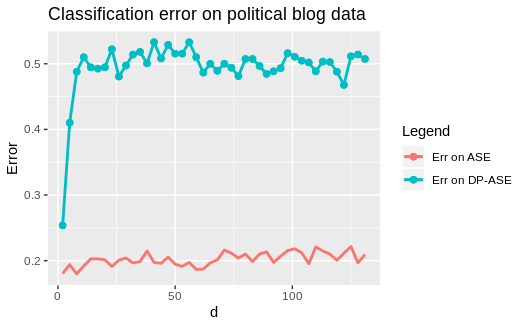}
}
\subfloat[]{
\includegraphics[width=0.26\textwidth]{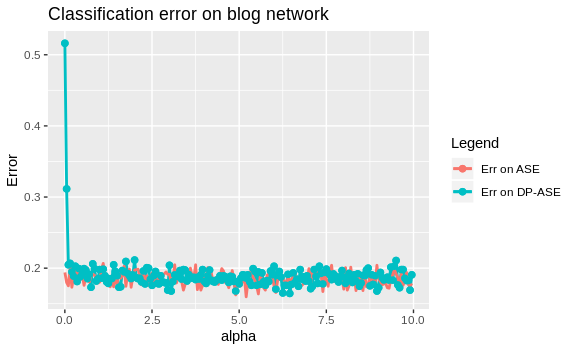}
}
\caption{\label{fig:poli}Experiment on political blog sphere. (a) Classification error by $k$NN$\circ$DP-ASE and $k$NN$\circ$ASE at embedding dimensions $d\in\{2, 5, 8, ..., 100\}$ with fixed $\alpha = 0.1$ and $\delta = 0.01$. (b) Privacy-utility tradeoff analysis. Utility-privacy tradeoff evaluation. As $\alpha$ increases, the error rate by $k$NN$\circ$DP-ASE decreases and approximates the error rate by $k$NN$\circ$ASE. For $\hat{d} = 2$, the lowest classification error by $k$NN$\circ$DP-ASE is 0.164 at best privacy parameter $\alpha= 6.451$. However we discover that at a privacy budget $\alpha$ as low as 0.251, $k$NN$\circ$DP-ASE has error of 0.189 very close to the classification error without privacy at 0.180.}
\end{figure}

The political blog sphere \citep{adamic2005political} contains 1490 blogs during the 2004 presidential election as vertices, and edges if the blogs are linked. Each blog is either liberal or conservative with label distribution at (0.508, 0.492). Hence the chance error is 0.492. Since the number of memberships does not necessarily equal to the correct embedding dimension in practice, we first vary the embedding dimension $d\in\{2, 5, 8, ..., 100\}$ at fixed $\alpha = 0.1$ and $\delta = 0.01$. As seen in Figure \ref{fig:poli} (a), the lowest classification error by $k$NN$\circ$DP-ASE is 0.25 at $\hat{d} = 2$, but it is higher than the classification error by $k$NN$\circ$ASE at 0.180. Such degradation is due to the noise added under differential privacy scheme.

Our next experiment is to examine the trade-off between privacy budget and utility by observing, at which values of $\alpha$, the error rate under privacy scheme is as accurate as the error rate without privacy. In practice, there exists limited guidance on appropriate values of $\alpha$ \cite{jayaraman2019relaxations} and thus we intend to understand such tradeoff via empirical evaluation. We fix $\delta = 0.01$, vary the privacy parameter $\alpha = 0.001:0.01:10$ and present the classification errors of $k$NN$\circ$DP-ASE and $k$NN$\circ$ASE at $\hat{d} =2$ in Fig  \ref{fig:poli} (b). As $\alpha$ increases, the error rate by $k$NN$\circ$DP-ASE decreases and approximates the error rate by $k$NN$\circ$ASE.  For $\hat{d} = 2$, the lowest classification error by $k$NN$\circ$DP-ASE is 0.164 at best privacy parameter $\alpha= 6.451$. However we discover that at a privacy budget $\alpha$ as low as 0.251, $k$NN$\circ$DP-ASE has error of 0.189 very close to the classification error without privacy at 0.180.



\subsubsection{Product co-purchasing network}

\begin{figure*}
\centering
\subfloat[]{
\includegraphics[width=0.3\textwidth]{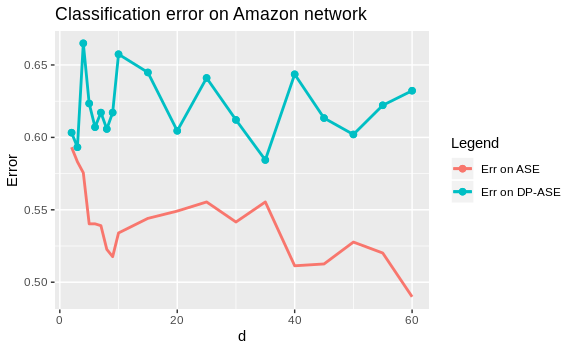}
}
\subfloat[]{
\includegraphics[width=0.3\textwidth]{{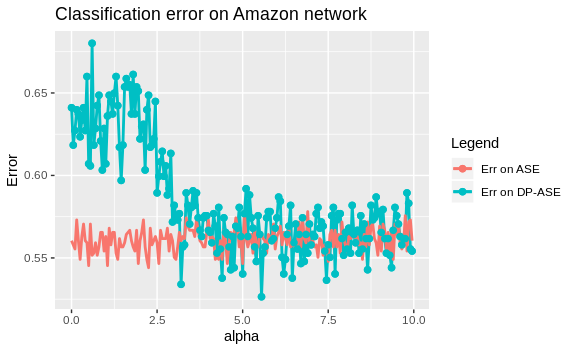}}
}
\subfloat[]{
\includegraphics[width=0.3\textwidth]{{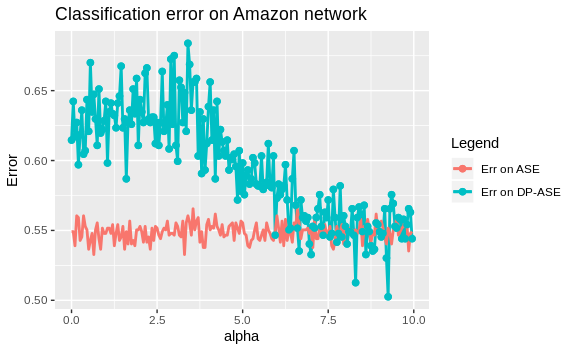}}
}
\caption{\label{fig:amazon}Experiment on product co-purchasing network. (a) Classification error by $k$NN$\circ$DP-ASE and $k$NN$\circ$ASE at embedding dimensions $d\in\{2, 3, ..., 60\}$ with fixed $\alpha = 0.1$ and $\delta = 0.01$. (b and c) Utility-privacy tradeoff evaluation. As $\alpha$ increases, the error rate by $k$NN$\circ$DP-ASE decreases and approximates the error rate by $k$NN$\circ$ASE. For $\hat{d} = 35$, the lowest classification error by $k$NN$\circ$DP-ASE is 0.52 at best privacy parameter $\alpha = 3.42$. For $\hat{d} = 60$, the lowest classification error by $k$NN$\circ$DP-ASE is 0.50 at best privacy parameter $\alpha = 9.25$.}
\end{figure*}

The Amazon product co-purchasing network \citep{yang2015defining} contains products as nodes with edge existence such that if a product $i$ is frequently co-purchased with product $j$, then an undirected edge exists between $i$ and $j$. The labels are product categories provided by Amazon. The original network contains 0.33M vertices with 0.92M edges. We select the six largest categories out of 5000 categories from the network. Since some products belong to multiple categories, we choose only one category to associate with each product and generate a subgraph of 794 nodes with label distribution of $(0.413, 0.393, 0.194)$. The classification chance error is 0.587. Since the number of memberships does not necessarily equal to the correct embedding dimension in practice, we first vary the embedding dimension $d \in \{2, 3, ..., 60\}$ at fixed $\alpha = 0.1$ and $\delta = 0.01$ and use 5-neareast neighbor classifier. We plot the loocv classification errors in Fig \ref{fig:amazon} (a) and find the best classification error under differential privacy scheme at 0.584 is only slightly lower than chance error. The contribution of noise under differential privacy is likely to cause the degradation in classification performance of  $k$NN$\circ$DP-ASE. The classification error without differential privacy is lowest at 0.489. We expect to see an improvement in classification efficacy, if better disjoint categories are selected to construct the subgraph or multi-label classification is performed. From the minimum error rates by $k$NN$\circ$DP-ASE and $k$NN$\circ$ASE respectively, 
we select two best embedding dimensions to be $\hat{d}_1 = 35$ and $\hat{d}_2 = 60$. 

Our next experiment is to examine the trade-off between privacy budget and utility by observing, at which values of $\alpha$, the error rate under privacy scheme is as accurate as the error rate without privacy. In practice, there exists limited guidance on appropriate values of $\alpha$ \cite{jayaraman2019relaxations} and thus we intend to understand such tradeoff via empirical evaluation. We fix $\delta = 0.01$, vary the privacy parameter $\alpha = 0.001:0.01:10$ and present the classification errors of $k$NN$\circ$DP-ASE and $k$NN$\circ$ASE at $\hat{d} \in \{\hat{d}_1, \hat{d}_2\}$ respectively in Fig \ref{fig:amazon} (b) and (c). As $\alpha$ increases, the error rate by $k$NN$\circ$DP-ASE decreases and approximates the error rate by $k$NN$\circ$ASE. For $\hat{d} = 35$, the lowest classification error by $k$NN$\circ$DP-ASE is 0.52 at best privacy parameter $\alpha = 3.42$. For $\hat{d} = 60$, the lowest classification error by $k$NN$\circ$DP-ASE is 0.50 at best privacy parameter $\alpha = 9.25$. 


\section{Conclusion}
In this paper we design a differentially private approximation to adjacency spectral embedding (DP-ASE) on stochastic blockmodels and investigate the theoretical and empirical performance of the proposed algorithm. In our numerical experiments, we demonstrate that as $n$ increases, for simulated stochastic blockmodels, a universally consistent classifier composed with DP-ASE can achieve the best error as ASE. From varying the privacy parameters in the simulation experiments, we observe that such mechanism can achieve the minimum possible error at desired privacy parameter levels, which can provide a tighter privacy bound. 
In practice, there has not been a principled guidance on the best privacy parameters to select while preserving inference accuracy. In our empirical analysis, we find that at reasonable privacy budget DP-ASE can achieve classification error as close as without preserving privacy. 
We continue our research with theoretical proofs on the consistency of DP-ASE and the universal consistency of a universally consistent classifier composed with DP-ASE. 


\nocite{langley00}

\bibliography{iclr2019_conference}

\begin{thebibliography}{18}
\providecommand{\natexlab}[1]{#1}
\providecommand{\url}[1]{\texttt{#1}}
\expandafter\ifx\csname urlstyle\endcsname\relax
  \providecommand{\doi}[1]{doi: #1}\else
  \providecommand{\doi}{doi: \begingroup \urlstyle{rm}\Url}\fi

\bibitem[Adamic \& Glance(2005)Adamic and Glance]{adamic2005political}
Adamic, L.~A. and Glance, N.
\newblock The political blogosphere and the 2004 us election: divided they
  blog.
\newblock In \emph{Proceedings of the 3rd international workshop on Link
  discovery}, pp.\  36--43. ACM, 2005.

\bibitem[Blum et~al.(2005)Blum, Dwork, McSherry, and Nissim]{blum2005practical}
Blum, A., Dwork, C., McSherry, F., and Nissim, K.
\newblock Practical privacy: the sulq framework.
\newblock In \emph{Proceedings of the twenty-fourth ACM SIGMOD-SIGACT-SIGART
  symposium on Principles of database systems}, pp.\  128--138. ACM, 2005.

\bibitem[Chaudhuri et~al.(2012)Chaudhuri, Sarwate, and
  Sinha]{chaudhuri2012near}
Chaudhuri, K., Sarwate, A., and Sinha, K.
\newblock Near-optimal differentially private principal components.
\newblock In \emph{Advances in Neural Information Processing Systems}, pp.\
  989--997, 2012.

\bibitem[Devroye et~al.(1996)Devroye, Gy{\"o}rfi, and
  Lugosi]{devroye1996probabilistic}
Devroye, L., Gy{\"o}rfi, L., and Lugosi, G.
\newblock \emph{A probabilistic theory of pattern recognition}, volume~31.
\newblock New York: Springer, 1996.

\bibitem[Dwork(2011)]{dwork2011differential}
Dwork, C.
\newblock Differential privacy.
\newblock \emph{Encyclopedia of Cryptography and Security}, pp.\  338--340,
  2011.

\bibitem[Fishkind et~al.(2013)Fishkind, Sussman, Tang, Vogelstein, and
  Priebe]{fishkind2013consistent}
Fishkind, D.~E., Sussman, D.~L., Tang, M., Vogelstein, J.~T., and Priebe, C.~E.
\newblock Consistent adjacency-spectral partitioning for the stochastic block
  model when the model parameters are unknown.
\newblock \emph{SIAM Journal on Matrix Analysis and Applications}, 34\penalty0
  (1):\penalty0 23--39, 2013.

\bibitem[Gower et~al.(2004)Gower, Dijksterhuis, et~al.]{gower2004procrustes}
Gower, J.~C., Dijksterhuis, G.~B., et~al.
\newblock \emph{Procrustes problems}, volume~30.
\newblock Oxford University Press on Demand, 2004.

\bibitem[Hoff et~al.(2002)Hoff, Raftery, and Handcock]{hoff2002latent}
Hoff, P.~D., Raftery, A.~E., and Handcock, M.~S.
\newblock Latent space approaches to social network analysis.
\newblock \emph{Journal of the American Statistical Association}, 97\penalty0
  (460):\penalty0 1090--1098, 2002.

\bibitem[Holland et~al.(1983{\natexlab{a}})Holland, Laskey, and
  Leinhardt]{holland1983stochastic}
Holland, P.~W., Laskey, K.~B., and Leinhardt, S.
\newblock Stochastic blockmodels: First steps.
\newblock \emph{Social networks}, 5\penalty0 (2):\penalty0 109--137,
  1983{\natexlab{a}}.

\bibitem[Holland et~al.(1983{\natexlab{b}})Holland, Laskey, and Leinhardt]{sbm}
Holland, P.~W., Laskey, K.~B., and Leinhardt, S.
\newblock Stochastic blockmodels: First steps.
\newblock \emph{Social networks}, 5\penalty0 (2):\penalty0 109--137,
  1983{\natexlab{b}}.

\bibitem[Jayaraman \& Evans(2019)Jayaraman and Evans]{jayaraman2019relaxations}
Jayaraman, B. and Evans, D.
\newblock When relaxations go bad:" differentially-private" machine learning.
\newblock \emph{arXiv preprint arXiv:1902.08874}, 2019.

\bibitem[Rohe et~al.(2011)Rohe, Chatterjee, and Yu]{rohe2011spectral}
Rohe, K., Chatterjee, S., and Yu, B.
\newblock Spectral clustering and the high-dimensional stochastic blockmodel.
\newblock \emph{The Annals of Statistics}, 39\penalty0 (4):\penalty0
  1878--1915, 2011.

\bibitem[Sussman et~al.(2012)Sussman, Tang, Fishkind, and Priebe]{stfp}
Sussman, D.~L., Tang, M., Fishkind, D.~E., and Priebe, C.~E.
\newblock A consistent adjacency spectral embedding for stochastic blockmodel
  graphs.
\newblock \emph{Journal of the American Statistical Association}, 107\penalty0
  (499):\penalty0 1119--1128, 2012.

\bibitem[Sussman et~al.(2014)Sussman, Tang, and Priebe]{sussman2014consistent}
Sussman, D.~L., Tang, M., and Priebe, C.~E.
\newblock Consistent latent position estimation and vertex classification for
  random dot product graphs.
\newblock \emph{IEEE Transactions on Pattern Analysis and Machine
  Intelligence}, 36\penalty0 (1):\penalty0 48--57, 2014.

\bibitem[Tang et~al.(2013)Tang, Sussman, and Priebe]{tang2013universally}
Tang, M., Sussman, D.~L., and Priebe, C.~E.
\newblock Universally consistent vertex classification for latent positions
  graphs.
\newblock \emph{The Annals of Statistics}, 41\penalty0 (3):\penalty0
  1406--1430, 2013.

\bibitem[Wang \& Wong(1987)Wang and Wong]{sbm2}
Wang, Y.~J. and Wong, G.~Y.
\newblock Stochastic blockmodels for directed graphs.
\newblock \emph{Journal of the American Statistical Association}, 82\penalty0
  (397):\penalty0 8--19, 1987.

\bibitem[Yang \& Leskovec(2015)Yang and Leskovec]{yang2015defining}
Yang, J. and Leskovec, J.
\newblock Defining and evaluating network communities based on ground-truth.
\newblock \emph{Knowledge and Information Systems}, 42\penalty0 (1):\penalty0
  181--213, 2015.

\bibitem[Young \& Scheinerman(2007)Young and Scheinerman]{young2007random}
Young, S.~J. and Scheinerman, E.~R.
\newblock Random dot product graph models for social networks.
\newblock In \emph{Algorithms and models for the web-graph}, pp.\  138--149.
  Springer, 2007.

\end{thebibliography}
\bibliographystyle{icml2019}


\end{document}